%% file: conformalgen_arxiv.tex
\documentclass{article}
\usepackage[preprint]{tmlr}
\input{_preamble}

\author{\name Mohammad Malekpouri \email m.malekpouri@iau.ir \\
  \addr Department of Computer Engineering, BA.C.\\
  Islamic Azad University, Bandar Abbas, Hormozgan, Iran
  \AND
  \name Somayeh Lotfi \email somayeh.lotfi@iau.ac.ir \\
  \addr Department of Computer Engineering, BA.C.\\
  Islamic Azad University, Bandar Abbas, Hormozgan, Iran}

\begin{document}
\maketitle
\input{abstract}

\input{body}
\bibliographystyle{tmlr}
\bibliography{references}
\end{document}

%% file: _preamble.tex
\usepackage{amsmath,amssymb,amsthm}
\usepackage{graphicx}
\usepackage{booktabs}
\usepackage{xcolor}
\usepackage[hyphens]{url}
\usepackage[colorlinks=true,allcolors=blue]{hyperref}

\newtheorem{theorem}{Theorem}
\newtheorem{proposition}{Proposition}

\newcommand{\FAR}{\mathrm{FAR}}
\newcommand{\ncal}{n_{\mathrm{cal}}}

\title{What Conformal Selection Can Certify in CRISPR Off-Target Design:\\
Benchmark Leakage, Free Baselines, and a Finite-Sample Cohort Requirement}

%% file: abstract.tex
\begin{abstract}
Conformal selection promises distribution-free, finite-sample control of the error rate of a
decision, and has been proposed as a safety layer for machine-learning CRISPR guide design. We ask
whether such a certificate is attainable from the validated off-target data that exist. The
constraint is arithmetic: a conformal p-value cannot fall below one divided by the
calibration cohort size plus one, so a Benjamini--Hochberg selection certifies a single candidate
only when that cohort is at least the candidate pool size divided by the tolerated error rate.
Public cohorts supply ten validated guides, whereas a pool of two thousand at a ten percent error
rate needs twenty thousand; the accepted set is consequently empty in over ninety-six percent of
splits. Two benchmarking findings explain why this went unnoticed. The standard
CIRCLE-seq file lists ten guides under eighty on-target strings, so grouping cross-validation on
the sequence column trains and tests on one guide; corrected, pair-level discrimination and
per-guide fidelity collapse. With that leakage removed, a training-free position-weight score and
a bare count of near-match genomic sites are not beaten by a trained oracle. We give the corrected
protocol, the baselines a new model must beat, and the cohort size a certified pipeline needs.
\end{abstract}

%% file: body.tex
\section{Introduction}
Machine-learning models now propose CRISPR/Cas9 guide RNAs faster than any
laboratory can validate them; a generative design pipeline of our own emits candidate spacers by
the million \citep{malekpouri2026crispgen}, and it is the immediate motivation for this work. That
makes the question of what a computational pipeline may \emph{promise} about an accepted guide a
practical one. A point prediction $\hat\mu(x)$ of off-target burden states nothing about how often
an accepted guide is in fact unsafe. Conformal prediction \citep{vovk2005,lei2018} supplies
distribution-free coverage under exchangeability, and conformal selection \citep{jin2023} converts
that into control of the error rate of a \emph{decision}: a Benjamini--Hochberg step
\citep{benjamini1995} applied to conformal $p$-values bounds the expected fraction of accepted
candidates that violate a design constraint. Because these methods wrap an arbitrary predictor
without retraining it, they are an attractive safety layer for genome editing.

Attractive, but not automatically applicable, and the obstacle is not statistical soundness. It is
a counting fact about the calibration sample, which we state as Proposition~\ref{prop:feas} and
take as the spine of this paper. Conformal $p$-values cannot fall below $1/(\ncal+1)$, so the
Benjamini--Hochberg threshold $qk/m$ is unreachable unless the calibration cohort is large relative
to the candidate pool; certifying a single candidate out of $m$ at level $q$ requires
$\ncal \ge m/q - 1$ validated calibration units. In off-target biology the calibration sample is
experimentally validated cleavage data, and the widely used CIRCLE-seq benchmark
\citep{tsai2017circleseq} contains ten guides. Certifying one guide from a generated pool of $2000$
at $q=0.10$ would require $19{,}999$. The bound is elementary --- it follows from the definitions
in two lines --- but it is, as far as we are aware, not stated in the conformal-selection
literature applied to genomics, and it is what explains the empty accepted sets we observe,
without any appeal to the quality of the underlying oracle. We note for completeness that
conformal $p$-values computed against a shared calibration set are dependent, so FDR control rests
on the positive-regression-dependence argument of \citet{jin2023} rather than on independence; the distribution-free alternative of \citet{benjamini2001} would inflate the threshold by a further
$\log m$ factor and tighten the cohort requirement rather than relax it.

That raises an obvious objection: if the data are two orders of magnitude short, why does this
literature report pair-level ROC-AUC above $0.9$ and successful certified selection? Our second
and third results answer it. The standard benchmark file silently encodes each guide many times,
so the natural grouping column for cross-validation is not the guide, and the resulting leakage is
large enough to account for the difference between a state-of-the-art-looking oracle and one that
does not beat a position-weight score published in 2013. And the quantity conformal selection
actually consumes --- a per-guide burden ranking --- is supplied essentially free by counting
enumerated near-match sites, and not at all by the trained oracle once the guide is unseen. We
then show that the comparison usually made in favour of a conformal rule, against a
calibration-tuned threshold, is a comparison at unequal yield.

Our conclusions are therefore largely cautionary, and we think more useful for it. The guarantees
are sound; what fails is their current data base, and both the shortfall and the remedy are
quantifiable.

\subsection{Contributions}
\begin{enumerate}
\item A feasibility bound for conformal selection, $\ncal \ge m/q - 1$ to certify one candidate
from a pool of $m$, with the cohort-size requirement it implies for CRISPR deployment and a design
sweep over cohort size and oracle fidelity (Sections~\ref{sec:method-feasibility}
and~\ref{sec:feasibility}).
\item A quantified account of representation-level leakage in a standard CRISPR off-target
benchmark, with the corrected grouping key and the effect size on three metrics
(Section~\ref{sec:leakage}).
\item A leak-free baseline comparison showing that free alignment counts and the classical
MIT/Hsu--Zhang score \citep{hsu2013} match or beat a trained pair-level oracle on the quantities
that matter downstream (Section~\ref{sec:baselines}).
\item A corrected evaluation of the usual uncertainty baselines and of cross-context transfer,
including a small-sample recalibration remedy that restores nominal coverage
(Section~\ref{sec:coverage}).
\end{enumerate}

\section{Materials and Methods}
\subsection{Data}
\textbf{Off-target.} CIRCLE-seq whole-dataset file \citep{tsai2017circleseq}: $584{,}949$
(guide, candidate-site) pairs, $7371$ validated cleavage events (positive rate $1.26\%$). The file
carries both a per-row aligned on-target string (\texttt{sgRNA\_seq}) and a canonical on-target
(\texttt{sgRNA\_type}); the latter takes exactly ten values and is the real unit of analysis
(Section~\ref{sec:leakage}). Two independent assays are used for transfer: SITE-Seq
\citep{cameron2017siteseq} (nine guides) and GUIDE-seq
\citep{tsai2015guideseq,listgarten2018elevation} ($22$ guides, $56$ validated events).
The scarcity is a property of the field rather than of this file. The other genome-wide
validation assays --- Digenome-seq \citep{kim2015digenome}, DISCOVER-seq
\citep{wienert2019discoverseq} and CHANGE-seq \citep{lazzarotto2020changeseq} --- each report
validated cleavage for at most a few tens of guides, so pooling every public source still leaves a
cohort far below what Proposition~\ref{prop:feas} will be shown to require.

\textbf{On-target efficacy.} Indel frequencies for $55{,}603$ wild-type SpCas9 guides from the
DeepHF panel \citep{wang2019deephf}, together with matched panels for engineered nucleases
(eSpCas9, SpCas9-HF1 \citep{kleinstiver2016hf}, Sniper-Cas9, xCas9, SpCas9-NG)
and three cell lines (HeLa, HCT116, HL-60), used for the coverage and transfer experiments.

\textbf{Genome.} GRCh38/hg38. Candidate off-target sites for novel guides are enumerated by an
exhaustive PAM scan: every NGG-adjacent $20$-mer on both strands, Hamming distance $\le 3$ over the
protospacer, no bulges, windows containing non-ACGT bases discarded. On chr22 this yields
$5{,}295{,}032$ candidate protospacers from $50{,}818{,}468$ bases. The scan is a bit-parallel
Hamming computation and needs no external aligner, but is equivalent to a Cas-OFFinder
\citep{bae2014casoffinder} $\mathrm{mm}\le3$ search restricted to the scanned footprint.

\subsection{Conformal Prediction and Conformal Selection}
We use split (inductive) conformal prediction throughout; \citet{angelopoulos2023conformal} is a recent survey, and \citet{bates2021rcps} extends the construction from coverage of a response to
distribution-free control of a general risk, which is the form the selection problem below takes.

Let $(X_i, Y_i)_{i=1}^{\ncal}$ be a calibration sample exchangeable with a test point $(X,Y)$, and
let $s(x,y)$ be a non-conformity score. We use the absolute residual and a directional variant;
conformalised quantile regression \citep{romano2019cqr} would give adaptive interval widths, and
Mondrian conformal prediction \citep{vovk2003mondrian} would give validity conditional on a
taxonomy such as the assay of origin --- both are orthogonal to the cohort-size obstruction
established below, which binds whatever the score. With
$\hat q_\alpha = s_{(\lceil (\ncal+1)(1-\alpha)\rceil)}$ the corresponding order statistic:

\begin{theorem}[Marginal coverage \citep{vovk2005,lei2018}]\label{thm:cov}
$C(x)=\{y: s(x,y)\le \hat q_\alpha\}$ satisfies $\Pr\{Y\in C(X)\}\ge 1-\alpha$, and if the scores
are almost surely distinct, $\Pr\{Y\in C(X)\}\le 1-\alpha+1/(\ncal+1)$.
\end{theorem}

The upper bound requires continuity. Our off-target scores are built from integer site counts and
therefore tie; we break ties conservatively (no randomisation), which preserves the lower bound and
voids the upper one.

For selection, a design specifies a tolerance $\tau$ on off-target burden,
$\mathrm{OK}(y) \iff y \le \tau$, and a monotone score $V(x,t) = t - \hat T(x)$ built from a
predictor $\hat T$ of the standardised margin. Conformal $p$-values
\begin{equation}
p_j = \frac{1+\sum_{i\le \ncal}\mathbb{1}[V(X_i,T_i)\le V(X_j,0)]}{\ncal+1}
\label{eq:pval}
\end{equation}
are super-uniform under the null, and BH at level $q$ controls the False Acceptance Rate
\citep{jin2023}:

\begin{proposition}[FAR control \citep{jin2023}]\label{prop:far}
Under exchangeability of the calibration and candidate pairs,
$\FAR(\widehat A) = \mathbb{E}[\,|\{j\in\widehat A: \neg\mathrm{OK}(Y_j)\}|
/ (|\widehat A|\vee 1)\,] \le q$.
\end{proposition}

\subsection{When Can a Conformal Selection Be Non-Empty?}
\label{sec:method-feasibility}
Proposition~\ref{prop:far} bounds an error rate but says nothing about yield, and at the cohort
sizes available in this field the yield constraint binds first.

\begin{proposition}[Feasibility]\label{prop:feas}
Conformal $p$-values satisfy $p_j \ge 1/(\ncal+1)$ for every candidate. BH at level $q$ over $m$
candidates rejects the $k$-th smallest only if $p_{(k)} \le qk/m$. Hence a non-empty accepted set
requires at least
\begin{equation}
k_{\min} \;=\; \left\lceil \frac{m}{q\,(\ncal+1)} \right\rceil
\label{eq:kmin}
\end{equation}
candidates attaining the minimal $p$-value, and certifying a single candidate $(k=1)$ requires
$\ncal \ge m/q - 1$.
\end{proposition}

\begin{proof}
By \eqref{eq:pval} the numerator is at least $1$, so $p_j \ge 1/(\ncal+1)$. BH rejects the $k$
smallest $p$-values for the largest $k$ with $p_{(k)} \le qk/m$; since no $p$-value is smaller than
$1/(\ncal+1)$, such a $k$ exists only if $1/(\ncal+1) \le qk/m$, that is
$k \ge m/(q(\ncal+1))$, which gives \eqref{eq:kmin}. Setting $k=1$ and rearranging gives
$\ncal \ge m/q - 1$.
\end{proof}

The consequence is practical: the calibration cohort must grow linearly in the candidate pool size
and inversely in the tolerated error rate. Certifying one guide out of a pool of $2000$ at
$q=0.10$ requires $19{,}999$ validated calibration guides; the public CIRCLE-seq cohort supplies
ten.

\subsection{Predictors and Baselines}
The pair-level oracle under audit encodes each (guide, site) pair by its per-position mismatch
pattern plus one-hot guide and off-target channels and the mismatch count ($181$ features) and is
fitted by gradient boosting. It is deliberately the simplest oracle sufficient to pose the
question. The standard comparators in this literature are deep pair-level models --- DeepCRISPR
\citep{chuai2018deepcrispr}, CRISPR-Net \citep{lin2020crisprnet}, the CNN of \citet{lin2018offtarget},
piCRISPR \citep{stortz2023picrispr} and DiPOff \citep{toufikuzzaman2024dipoff}, surveyed by \citet{sherkatghanad2023review} --- but each is reported on its own split and none we examined
states a grouping unit coarser than the aligned on-target string, so Section~\ref{sec:leakage}
bears on them as a class rather than describing a defect of one implementation. We therefore treat
published pair-level figures as not comparable to ours rather than as a benchmark we beat. We
compare against: (i) the mismatch count; (ii) the MIT/Hsu--Zhang
position-weighted score \citep{hsu2013}; (iii) a transparent position-substitution logistic model
($20$ positions $\times$ $16$ base pairings $=320$ features), which is a fully specified stand-in
for a learned CFD-type model; and, for the per-guide burden that selection actually consumes,
(iv) the number of enumerated candidate sites and (v) the number of $\mathrm{mm}\le3$ sites, both
free by-products of the alignment step. Baselines (i), (iv) and (v) are exactly the quantities a
rule-based genome-wide screen already returns \citep{haeussler2016,concordet2018}, so a trained
oracle that does not beat them is not buying anything a designer does not have. CFD itself \citep{doench2016} is not re-run here because its
penalty matrix is distributed only as a binary pickle; published CFD numbers on these benchmarks
should be tabulated from their sources rather than re-estimated. The MIT/Hsu--Zhang weight vector
we use is the one distributed with CRISPRseek (Bioconductor, \texttt{getOfftargetScore}, default
\texttt{weights}, attributed there to \citet{hsu2013}); the three-term formula is as reproduced in
the patent literature deriving from that work.

\subsection{Evaluation Protocol}
All cross-validation and all calibration/test splits are grouped on \texttt{sgRNA\_type}, that is,
on the real guide. Discrimination is reported as average precision alongside ROC-AUC, since the
positive rate is $1.26\%$, and both are reported pooled and stratified within guide, because the
pooled figure confounds ranking sites within a guide with ranking guides against each other.
Per-guide fidelity carries a $95\%$ confidence interval from bootstrapping \emph{guides}, the
independent unit, which at $n=10$ is necessarily wide --- that width is information, not noise to
be hidden. Design tolerances $\tau$ are fixed a priori in absolute units (validated cleavage sites)
rather than taken as a quantile of the labels being evaluated.

\section{Results}
\subsection{Certified Selection Is Feasible Only at Cohort Sizes the Field Does Not Have}
\label{sec:feasibility}
Applying conformal selection to the ten-guide cohort at the real-guide level, with a
pre-registered absolute tolerance, the accepted set is empty in at least $96.7\%$ of
calibration/test splits at every tolerance and both levels $q\in\{0.10,0.20\}$, with mean power
below $0.04$. Proposition~\ref{prop:feas} explains this without reference to the oracle: with
$\ncal \le 7$ the smallest attainable $p$-value is $\ge 0.125$, while BH requires
$p_{(k)} \le qk/m$.

\begin{figure*}[!t]
\centering
\includegraphics[width=\textwidth]{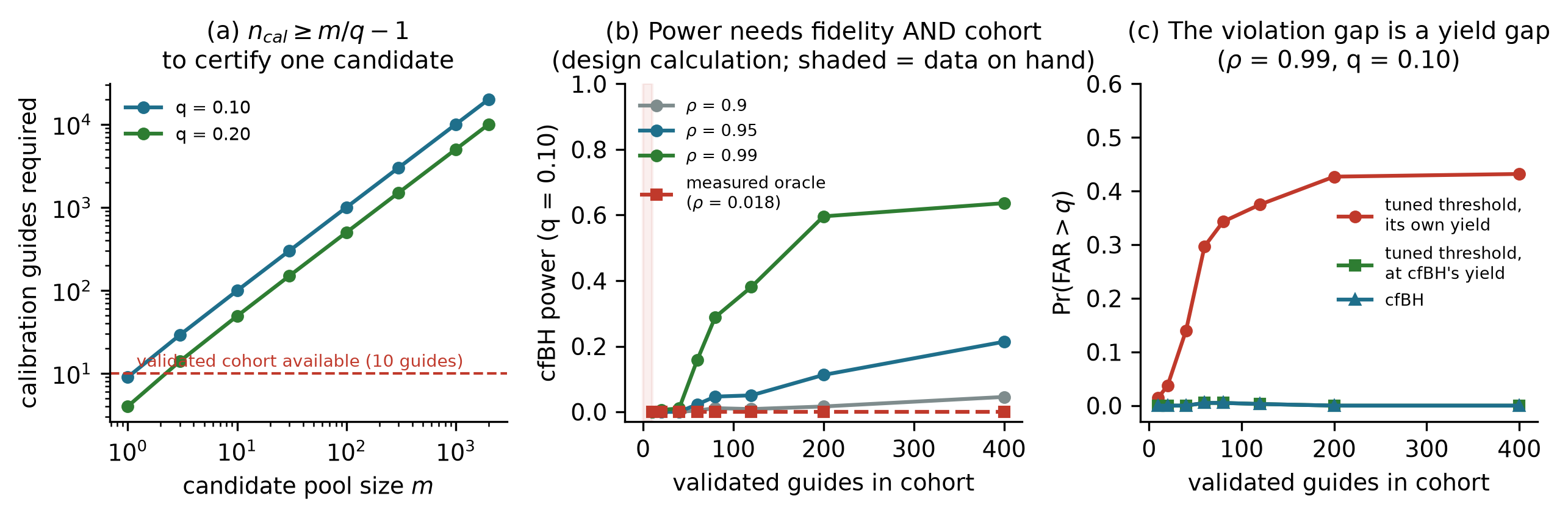}
\caption{(a) The feasibility bound of Proposition~\ref{prop:feas}; (b) power as a function of
cohort size and oracle fidelity; (c) the tuned-threshold violation gap closes at matched yield.}
\label{fig:feas}
\end{figure*}

Fig.~\ref{fig:feas}(b) turns the bound into a design calculation, anchored on the measured
per-guide burden distribution (median $381$ validated sites, range $169$--$2513$) and sweeping
cohort size against oracle fidelity. At $q = 0.10$, mean power reaches $0.5$ only at
$\rho \approx 0.99$ \emph{and} $n \approx 200$ validated guides; at the fidelity of the best free
predictor ($\rho \approx 0.95$) power is $0.21$ even at $n = 400$; at the measured fidelity of the
trained oracle ($\rho = 0.018$, Section~\ref{sec:leakage}) it is zero everywhere. At the more
permissive $q = 0.20$, $n \approx 80$ suffices when $\rho \approx 0.99$. These are simulations from
a measured distribution, not data, and we label them as such --- but the conclusion is robust to
that caveat, because Proposition~\ref{prop:feas} is an identity.

\subsubsection{The certificate's value, measured at matched yield}
Conformal selection is usually defended against a calibration-tuned threshold, which is reported
to violate the target on a large fraction of deployments. We reproduce that gap, and show it is a
yield gap (Fig.~\ref{fig:feas}(c)). At its own operating point the tuned threshold accepts roughly
twice as much and violates $\FAR \le q$ on $37$--$47\%$ of splits; constrained to accept exactly
what conformal selection accepted, its FAR is $0.001$, identical to conformal's, and its violation
rate falls to $0.01$. The certificate therefore does not make a given accepted set safer; it
selects a conservative operating point automatically and with a finite-sample justification, which
is a real but much narrower benefit.

\subsection{The Benchmark Encodes Ten Guides as Eighty}\label{sec:leakage}
The CIRCLE-seq file lists each guide under several on-target strings. Stripping the gap character
from \texttt{sgRNA\_seq} yields $80$ distinct strings and $800$ raw strings; the canonical
\texttt{sgRNA\_type} column yields exactly ten. The $80$ are ten protospacers $\times$ eight
representations: four PAM variants (NAGG/NCGG/NGGG/NTGG) crossed with two alignment hypotheses, a
mismatch-only alignment and a bulge alignment (the leading \texttt{-} is an alignment pad at index
$0$; \texttt{\_} is the bulge marker). Because feature extraction truncates the guide to its first
$20$ nucleotides, the guide-side features of these eight representations are bit-identical, so
grouping cross-validation on \texttt{sgRNA\_seq} places the same guide, with the same feature
vector, on both sides of a nominally guide-disjoint split. The failure is an instance of the
leakage taxonomy of \citet{kapoor2023leakage}, who find leakage affecting at
least $294$ papers across $17$ disciplines; the variant here is theirs in which train and test
samples are not independent, arrived at through a grouping key finer than the biological unit.

\begin{figure*}[!t]
\centering
\includegraphics[width=\textwidth]{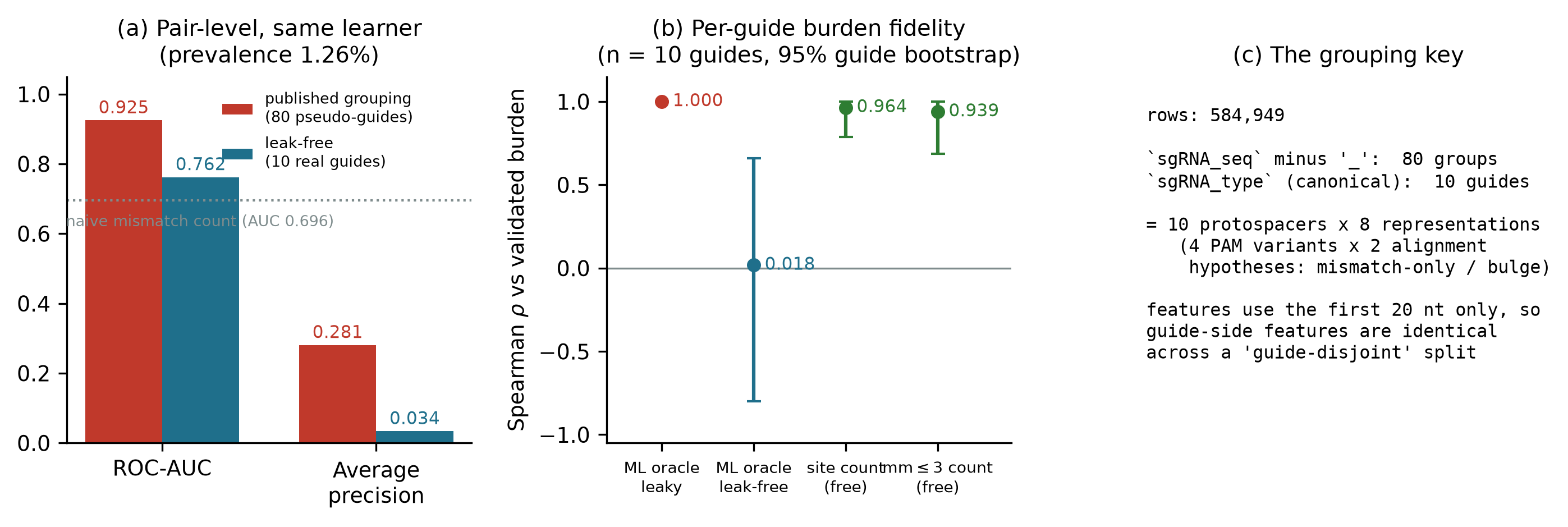}
\caption{Representation-level leakage in a standard CRISPR off-target benchmark. Same learner,
same features, same hyperparameters; only the grouping key differs.}
\label{fig:leak}
\end{figure*}

Fig.~\ref{fig:leak} quantifies the consequence with the learner, the features and the
hyperparameters held fixed. Pair-level ROC-AUC falls from $0.925$ to $0.762$ and average precision
from $0.281$ to $0.034$ against a positive rate of $1.26\%$ --- that is, the leak-free average
precision is close to the mismatch-count baseline ($0.026$). The per-guide burden fidelity on which
conformal selection depends falls from $\rho=1.000$, a value that is itself diagnostic of
memorisation, to $\rho = 0.018$, with a guide-bootstrap interval of $[-0.80, 0.66]$. The trained
oracle, in other words, carries no information about the off-target burden of a guide it has not
seen --- and it is exactly that fidelity which Fig.~\ref{fig:feas}(b) requires to be near $0.99$
before certified selection has any power.

\subsection{Free and Classical Baselines Match or Beat the Trained Oracle}\label{sec:baselines}
\begin{figure*}[!t]
\centering
\includegraphics[width=\textwidth]{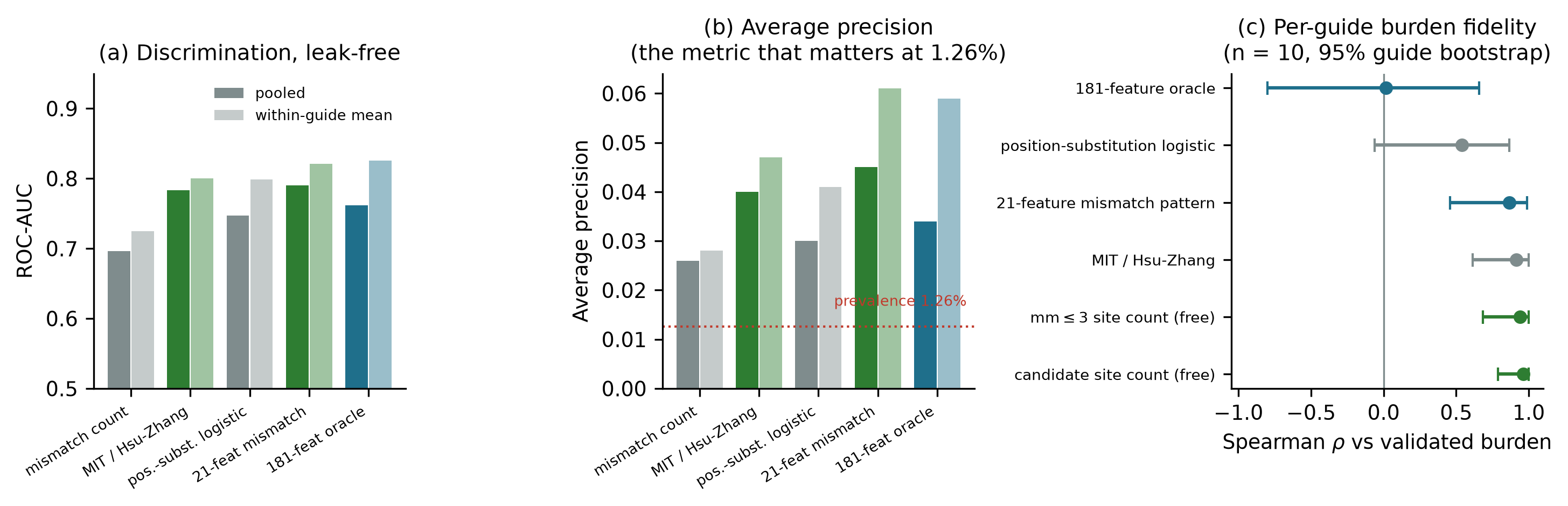}
\caption{Leak-free comparison. (a,b) pair-level discrimination, pooled and within guide;
(c) per-guide burden fidelity, the quantity conformal selection consumes.}
\label{fig:base}
\end{figure*}

Fig.~\ref{fig:base} and Table~\ref{tab:base} place the leak-free oracle beside baselines that cost
nothing to compute. At the pair level the MIT/Hsu--Zhang score --- twenty published weights, no
training --- attains AUC $0.783$ and AP $0.040$, against $0.762$ and $0.034$ for the $181$-feature
oracle. The oracle's one advantage is within-guide: it ranks sites inside a guide slightly better
than the alternatives (AUC $0.826$ against $0.800$; AP $0.059$ against $0.047$), which is the
regime a per-site triage tool operates in.

The informative contrast is with the \emph{same} learner stripped of its guide and off-target
identity channels. A $21$-feature model on the mismatch pattern alone attains the best pair-level
figures in the table (AUC $0.790$, AP $0.045$) and retains per-guide burden fidelity
$\rho = 0.867$ $[0.46, 0.99]$, where the $181$-feature model collapses to $\rho = 0.018$
$[-0.80, 0.66]$. The one-hot identity channels let the model memorise guides it has seen and leave
it with nothing transferable for guides it has not; removing them improves every metric that
matters.

For the per-guide burden that conformal selection consumes, the ordering is nonetheless stark:
counting the enumerated candidate sites gives $\rho = 0.964$ $[0.787, 1.000]$ and counting
$\mathrm{mm}\le3$ sites $\rho = 0.939$ $[0.686, 1.000]$, better than any trained model here. Total
off-target burden is, to first order, a property of how many near-matches a guide's spacer has in
the genome: an alignment statistic, not a learned one. With ten guides these correlations carry
wide intervals and are unstable across fold schemes --- the $181$-feature model's $\rho$ moves
between $0.018$ and $0.273$ depending on whether folds hold out one guide or two --- which is
itself a reason to report the interval rather than the point estimate.

\begin{table*}[!t]
\renewcommand{\arraystretch}{1.25}
\caption{Off-target scoring, leak-free (grouped on the real guide; $584{,}949$ pairs, ten guides,
positive rate $1.26\%$). Burden $\rho$ is the Spearman correlation with validated per-guide burden,
with a $95\%$ guide-bootstrap interval.}
\label{tab:base}
\centering
\small
\begin{tabular}{@{}lcccc@{}}
\toprule
& \multicolumn{2}{c}{pooled} & within-guide & per-guide \\
\cmidrule(lr){2-3}
Scorer & ROC-AUC & AP & ROC-AUC & burden $\rho$ \\
\midrule
mismatch count                 & $0.696$ & $0.026$ & $0.725$ & --- \\
MIT/Hsu--Zhang \citep{hsu2013}  & $0.783$ & $0.040$ & $0.800$ & $0.915$ $[0.62,1.00]$ \\
pos.-subst. logistic           & $0.747$ & $0.030$ & $0.799$ & $0.539$ $[-0.06,0.87]$ \\
$21$-feat. mismatch pattern    & $\mathbf{0.790}$ & $\mathbf{0.045}$ & $0.821$ & $0.867$ $[0.46,0.99]$ \\
$181$-feat. oracle (audited)   & $0.762$ & $0.034$ & $\mathbf{0.826}$ & $0.018$ $[-0.80,0.66]$ \\
\midrule
candidate-site count (free)    & --- & --- & --- & $\mathbf{0.964}$ $[0.79,1.00]$ \\
$\mathrm{mm}\le3$ count (free) & --- & --- & --- & $0.939$ $[0.69,1.00]$ \\
\bottomrule
\end{tabular}
\end{table*}

\subsection{Coverage Holds in Distribution; the Usual Uncertainty Baselines Are Not Broken}
\label{sec:coverage}
The efficacy axis is included because it is where published on-target predictors are strongest
\citep{kim2019deepspcas9,xiang2021crispron} and therefore where a conformal wrapper has the best
prospect of certifying something; the off-target axis is the harder case.
On $55{,}603$ wild-type efficacy measurements, split-conformal coverage matches its nominal level:
$0.899$ $[0.896, 0.901]$ for the absolute-residual score and $0.900$ $[0.898, 0.903]$ for the
directional score at $1-\alpha = 0.90$. This is the empirical behaviour Theorem~\ref{thm:cov}
predicts, and we state it as such rather than as exactness.

\begin{figure}[!t]
\centering
\includegraphics[width=0.6\textwidth]{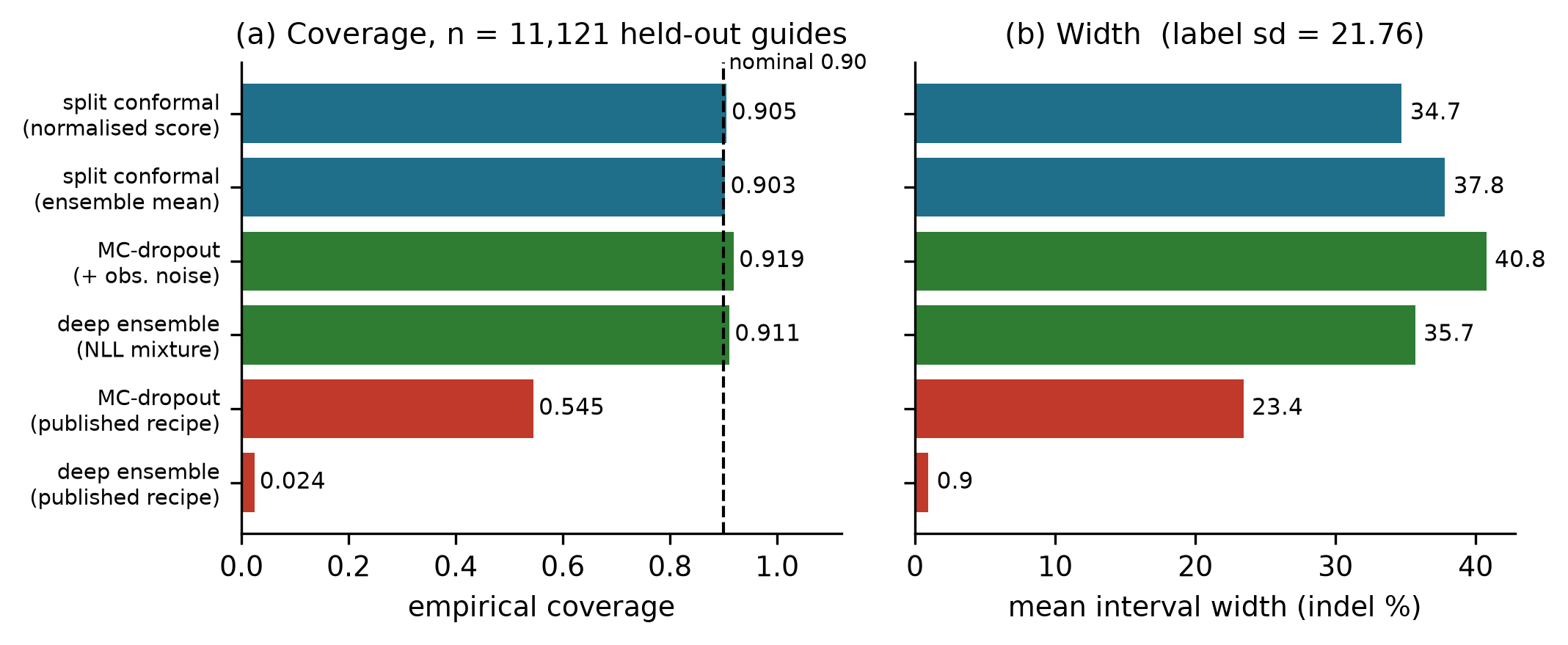}
\caption{Uncertainty baselines on $11{,}121$ held-out guides. The published recipe is reproduced
verbatim; the corrected arms add the predictive-noise term their source papers specify.}
\label{fig:uq}
\end{figure}

A frequent claim in this literature is that conformalisation repairs the miscalibration of standard
deep-learning uncertainty methods --- a claim whose premise, that modern networks are badly
miscalibrated out of the box \citep{guo2017calibration}, is itself about classification confidence
rather than regression intervals. It does not, once those methods are implemented as specified
(Fig.~\ref{fig:uq}). Forming a deep-ensemble interval as
$\hat\mu \pm z\,\mathrm{sd}(\text{member means})$ uses the epistemic spread only and omits the
predictive-noise term; it yields coverage $0.024$ at nominal $0.90$ with a mean width of $0.95$ on
a $0$--$100$ scale whose label standard deviation is $21.8$, a number forced by arithmetic rather
than by the method. With Gaussian negative-log-likelihood heads and the mixture predictive variance
of \citep{lakshminarayanan2017}, the same five-member ensemble covers at $0.911$; MC-dropout with
the observation-noise term of \citep{gal2016dropout} covers at $0.919$. Split conformal on the same
predictors covers at $0.903$ and, with a variance-normalised score, at $0.905$ with the narrowest
interval of all arms. The honest statement is therefore that conformal needs no distributional
assumption and gives competitive width, not that it repairs a broken competitor.

\subsection{Transfer Fails, and a Small Labelled Cohort Repairs It}
\begin{figure*}[!t]
\centering
\includegraphics[width=\textwidth]{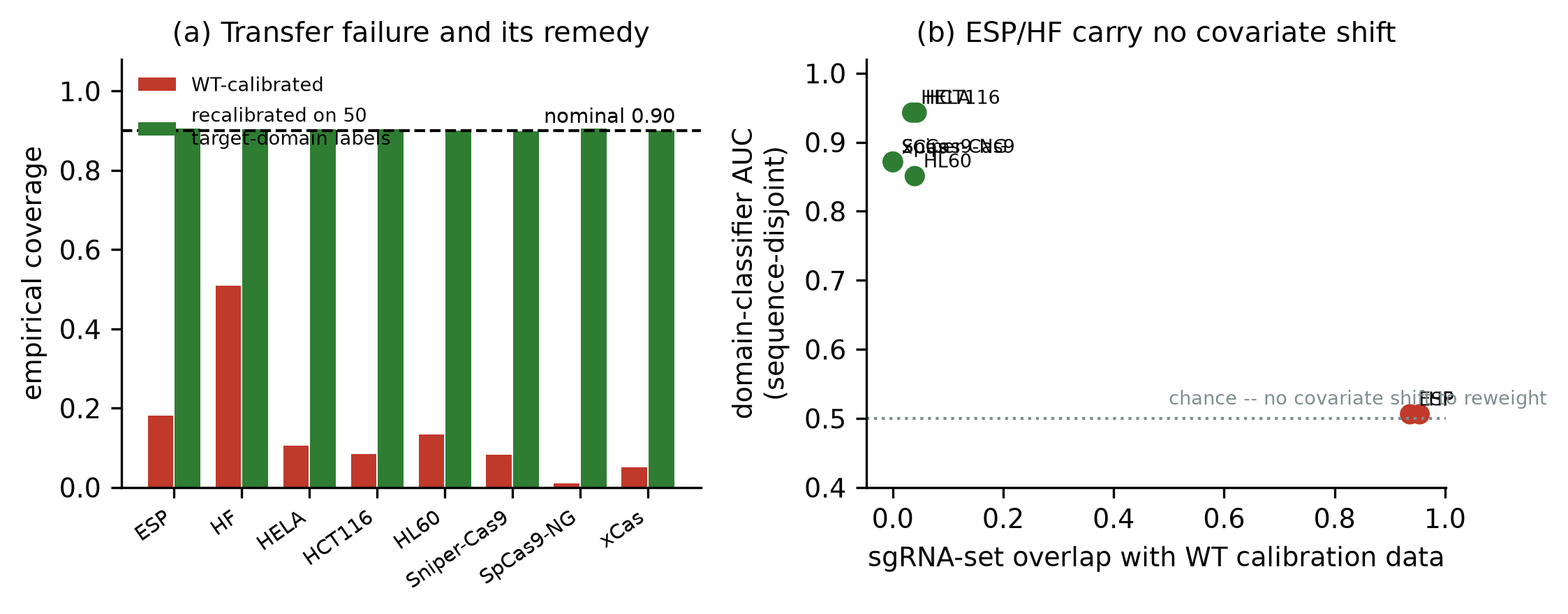}
\caption{Cross-context transfer. (a) coverage before and after recalibration on $50$
target-domain labels; (b) two panels carry no covariate shift to reweight.}
\label{fig:trans}
\end{figure*}

A wild-type-calibrated conformal predictor under-covers badly on engineered nucleases and other
cell lines: $0.182$ (eSpCas9), $0.509$ (SpCas9-HF1), $0.105$ (HeLa), $0.085$ (HCT116), $0.133$
(HL-60), and $0.010$--$0.083$ for Sniper-Cas9, SpCas9-NG and xCas9, all at nominal $0.90$
(Fig.~\ref{fig:trans}).

Two qualifications matter. First, covariate-shift weighting \citep{tibshirani2019}, and the
weighted-quantile construction that relaxes exchangeability more generally
\citep{barber2023beyond}, cannot be evaluated on the two largest panels at all: eSpCas9 and SpCas9-HF1 share $93.7\%$ and $95.4\%$ of
their guide sets with the wild-type calibration data, so the density ratio is $1$ and weighted
conformal reduces to split conformal. A domain classifier reports chance discrimination
(AUC $0.506$) once its split is sequence-disjoint; a higher apparent accuracy reflects duplicated
sequences, not a detectable shift. These panels are matched-input, shifted-label --- a concept
shift, which no covariate reweighting addresses. The genuinely shifted panels are the cell lines
($3.5$--$4.3\%$ overlap, domain AUC $0.85$--$0.94$).

Second, the effective remedy is simple and cheap: recalibrating the conformal quantile on $50$
target-domain labels restores coverage to $0.899$--$0.906$ in \emph{every} panel, shifted inputs or
not. Conformal prediction transfers when it is allowed a small in-domain calibration sample, and
this --- not reweighting --- is the deployment recommendation the data support.

\subsection{An End-to-End Generate--Align--Score--Certify Loop}
To test the pipeline rather than its parts, we trained a character-level autoregressive generator
on $55{,}603$ wild-type spacers --- a deliberately plain stand-in for the heavier generative models
this line of work has used before \citep{malekpouri2026crispgen}, since the certification step
treats the generator as opaque --- sampled $2000$ novel candidates ($100\%$ novel), enumerated
their off-target sites against chr22 by exhaustive PAM scan, scored every enumerated site with the
CIRCLE-trained per-site oracle (trained on the $332{,}070$ mismatch-only pairs carrying $7031$
validated events), and ran the certification step calibrated on the ten validated guides. The loop
runs; the accepted set is empty. Proposition~\ref{prop:feas} gives the reason --- certifying from a
pool of $2000$ at $q = 0.10$ with $\ncal = 10$ would require $1819$ candidates at the minimal
$p$-value --- and a second obstacle compounds it: the calibration cohort's truth is a genome-wide
validated count while the candidates' scores come from a chr22-restricted scan, and the public file
carries no site coordinates with which to put both on one footprint. Exchangeability between
calibration and candidate pairs, not oracle quality, is what blocks the loop.

\section{Discussion}
Three lessons generalise beyond this dataset.

\textbf{Distribution-free does not mean data-free.} Proposition~\ref{prop:feas} is elementary but
binds hard: the calibration cohort must scale with the candidate pool. For a shortlist of ten
candidates at $q=0.10$ one needs about a hundred validated guides; for a generative pool of
thousands, tens of thousands. The realistic near-term deployment of conformal selection in this
field is therefore triage of a short shortlist, with an in-assay validated cohort of order $10^2$
guides --- not the filtering of large generated libraries, which is how it is usually proposed. A
corollary for practice is that reporting the realised yield alongside the error rate is not
optional: an empty or near-empty accepted set satisfies every FAR bound trivially.

\textbf{Group on the biological unit, and check what the file encodes.} Off-target benchmark files
routinely list one guide under many alignment representations. Any grouping key other than the
canonical on-target silently permits within-guide leakage, and the effect here was large enough to
move a pair-level AUC from $0.76$ to $0.93$ and a per-guide correlation from $0.02$ to $1.00$. We
recommend that studies on these files state their grouping column explicitly, report average
precision next to AUC at the stated prevalence, and report within-guide as well as pooled
discrimination. The grouped-shift benchmarking conventions of WILDS \citep{koh2021wilds} are the
closest existing template, and guide-design evaluation has begun to move the same way
\citep{kim2026guideeval}.

That recommendation includes our own earlier work. The generative pipeline cited above
\citep{malekpouri2026crispgen} trains its off-target discriminator on a pooled multi-assay corpus
partitioned at the level of the individual (guide, site) pair rather than the guide; $99.9\%$ of
its validation rows carry an sgRNA that also appears in training. Retraining that architecture
unchanged and altering nothing but the split, validation Pearson $R$ falls from $0.811$ --- which
reproduces the published figure --- to $0.464$, and a correction is in preparation. We mention it
because the failure mode is easy to introduce and hard to notice: a random split over rows is the
default in every standard training script, it raises no error, and it returns a number that looks
like success. In this domain the row is not the independent unit --- the guide is, and public
corpora contain three orders of magnitude fewer guides than rows.

A related trap appears in the same place. That corpus pools six assays whose labels are on
incompatible scales (read counts, read fractions, Day21-ETP, indel rates), and a Pearson
correlation computed over the pool rewards separating assays as much as ranking sites within one.
At the guide-disjoint optimum the pooled $R$ of $0.464$ exceeds the within-assay $R$ of
\emph{every} constituent assay, the largest of which is $0.104$ --- an instance of Simpson's
paradox, and a second inflation independent of the first. Nor is it repaired by the obvious
remedies: normalising scores within each assay before pooling, or fitting one model per assay,
leaves within-assay Spearman at $0.00$--$0.15$ everywhere. Where labels come from heterogeneous
assays, the quantity to report is within-assay, and a pooled correlation that exceeds every
constituent one should be read as a warning rather than a result.

\textbf{Compare against what is free.} Per-guide off-target burden is dominated by the number of
near-match sites in the genome, which the alignment step already produces. A learned pair-level
model should be asked to beat that count, and a per-site model should be asked to beat
MIT/Hsu--Zhang and CFD, before it is described as validated. Relatedly, guide-identity features are
what make leakage profitable: dropping them cost nothing here and improved every metric, so we
would advise against one-hot guide channels in any model meant to generalise to unseen guides.

\subsection{What Survives of the Conformal Programme}
The guarantees themselves are not in question: in-distribution coverage behaved as
Theorem~\ref{thm:cov} predicts at $55$k scale, FAR control held in every cell we examined, and the
fail-safe behaviour of an empty accepted set is genuinely preferable to an uncontrolled one. What
does not survive is the claim that a validated oracle plus conformal selection currently delivers
useful off-target safety selection, and the claim that conformalisation is needed to repair
standard uncertainty quantification.

\subsection{Limitations}
The validated cohort is ten guides; every per-guide statement here carries a wide interval, which
we report rather than suppress. The genome scan covers chr22 and excludes bulges, so the enumerated
burdens are a lower bound; given that bulge-annotated rows carry the majority of validated events
in this file, a bulge-aware scan is the natural extension. The cohort-size sweep beyond $n=10$ is
simulated from the measured burden distribution. The MIT/Hsu--Zhang weights are those distributed
with CRISPRseek and attributed there to \citet{hsu2013}; we were unable to consult the original
supplementary table, so the vector rests on agreement between independent implementations. Finally,
no cohort carries both validated efficacy and validated off-target labels, so joint
multi-objective certification remains untested.

\section{Conclusion}
Conformal selection gives CRISPR guide design a finite-sample certificate, and the certificate is
sound. Whether it is \emph{attainable} is a question about data, and at present the answer in
off-target design is no: the validated cohorts are two orders of magnitude smaller than the
feasibility bound requires, the trained oracle that was supposed to drive the selection carries no
per-guide signal once leakage is removed, and a free alignment count does the job it was meant to
do. We offer the corrected benchmarking protocol, the baselines a new off-target model must beat,
and the cohort-size rule that tells a laboratory how much validated data a certified pipeline would
actually need.

\section*{Data and Code Availability}
All code, the audit scripts that produce every number and figure in this paper, and the
machine-readable results are released under the MIT license at
\url{https://github.com/malekpouri/conformal-genomics}. The script \texttt{reproduce.sh}
regenerates the pipeline and \texttt{audit/} regenerates this paper's analyses on CPU. The
CIRCLE-seq, SITE-Seq, GUIDE-seq and indel-frequency datasets are public and cited above; GRCh38 is
available from UCSC.

\section*{Author Contributions}
M.M.: conceptualisation, methodology, software, analysis, writing. S.L.: supervision, methodology,
review and editing.

\section*{Conflict of Interest}
The authors declare no competing interests.